\pdfoutput=1

\documentclass[11pt]{article}

\usepackage[final]{acl}

\usepackage{times}
\usepackage{latexsym}

\usepackage[T1]{fontenc}

\usepackage[utf8]{inputenc}

\usepackage{microtype}

\usepackage{inconsolata}
\usepackage{booktabs}

\usepackage{graphicx}

\usepackage{hyperref}
\usepackage{url}
\usepackage{multirow, multicol}
\usepackage{diagbox}

\usepackage{ctable}
\usepackage{tabularx}
\usepackage{xspace}
\usepackage{amsmath}
\usepackage{float}
\usepackage{xcolor}
\usepackage{soul}
\usepackage{tablefootnote}
\usepackage{colortbl}
\usepackage{pifont}
\usepackage{amssymb}
\usepackage{enumitem}
\usepackage{listings}
\usepackage{listings-ext}
\usepackage{longtable}

\newcommand{\system}[0]{\textsc{Spaghetti}\xspace}

\newcommand{\systemEmoji}{\includegraphics[height=1.1em,trim=5em 7em 0 0]{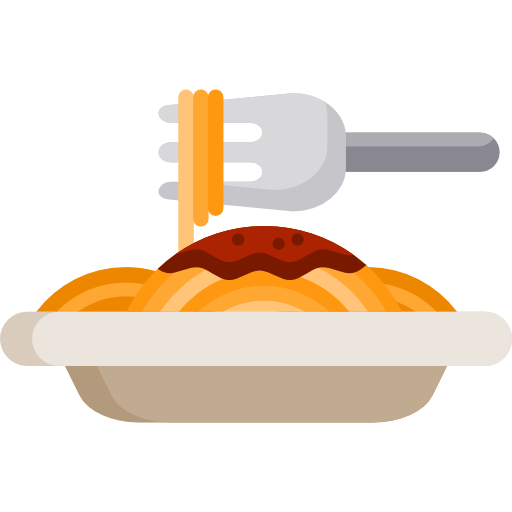}}
\newcommand{\systemWithEmoji}{\systemEmoji\xspace\system}

\newcommand{\wikichat}[0]{WikiChat\xspace}

\newcommand{\compmix}[0]{\textsc{CompMix}\xspace}
\newcommand{\convmix}[0]{\textsc{ConvMix}\xspace}

\definecolor{punct}{rgb}{0.5,0,0}
\definecolor{delim}{rgb}{0,0.5,0}
\definecolor{key}{rgb}{0,0,0.5}
\definecolor{num}{rgb}{0.5,0.5,0.5}     
\definecolor{background}{rgb}{0.95,0.95,0.95} 

\lstdefinelanguage{json}{
    basicstyle=\footnotesize\ttfamily,
    numbers=left,
    numberstyle=\scriptsize,
    stepnumber=1,
    numbersep=8pt,
    showstringspaces=false,
    breaklines=true,
    frame=lines,
    backgroundcolor=\color{background},
    literate=
     *{0}{{{\color{num}0}}}{1}
      {1}{{{\color{num}1}}}{1}
      {2}{{{\color{num}2}}}{1}
      {3}{{{\color{num}3}}}{1}
      {4}{{{\color{num}4}}}{1}
      {5}{{{\color{num}5}}}{1}
      {6}{{{\color{num}6}}}{1}
      {7}{{{\color{num}7}}}{1}
      {8}{{{\color{num}8}}}{1}
      {9}{{{\color{num}9}}}{1}
      {:}{{{\color{punct}{:}}}}{1}
      {,}{{{\color{punct}{,}}}}{1}
      {\{}{{{\color{delim}{\{}}}}{1}
      {\}}{{{\color{delim}{\}}}}}{1}
      {[}{{{\color{delim}{[}}}}{1}
      {]}{{{\color{delim}{]}}}}{1},
}

\title{\systemWithEmoji: Open-Domain Question Answering \\from Heterogeneous Data Sources with Retrieval and Semantic Parsing}

\author{Heidi C. Zhang \quad Sina J. Semnani \quad Farhad Ghassemi \quad Jialiang Xu \\ {\bf Shicheng Liu} \quad {\bf Monica S. Lam} \\
Computer Science Department, Stanford University \\ Stanford, CA \\
        \texttt{\{chenyuz, sinaj, farhadg, xjl, shicheng, lam\}@cs.stanford.edu}}

\begin{document}
\maketitle
\begin{abstract}
We introduce \systemWithEmoji: \textbf{S}emantic \textbf{P}arsing \textbf{A}ugmented \textbf{G}eneration for \textbf{H}ybrid \textbf{E}nglish information from \textbf{T}ext \textbf{T}ables and \textbf{I}nfoboxes, a hybrid question-answering (QA) pipeline that utilizes information from heterogeneous knowledge sources, including knowledge base, text, tables, and infoboxes. Our LLM-augmented approach achieves state-of-the-art performance on the \compmix dataset, the most comprehensive heterogeneous open-domain QA dataset, with 56.5\% exact match (EM) rate. More importantly, manual analysis on a sample of the dataset suggests that \system is more than 90\% accurate, indicating that EM is no longer suitable for assessing the capabilities of QA systems today.\footnote{Code is available at \href{https://github.com/stanford-oval/WikiChat}{https://github.com/stanford-oval/WikiChat}.}
\end{abstract}

\section{Introduction}



Open-domain question answering (QA) grounded in knowledge corpora has long been an active topic of research in natural language processing~\citep{chen-etal-2017-reading, Wang_Yu_Guo_Wang_Klinger_Zhang_Chang_Tesauro_Zhou_Jiang_2018, lee-etal-2019-latent, Asai2020Learning, izacard-grave-2021-leveraging, khattab-etal-2021-relevance, asai-etal-2022-evidentiality}.
With the rise of LLMs, new state of the art has been established with QA {\em separately} on free-text documents~\citep{semnani-etal-2023-wikichat, jiang-etal-2023-active, gao-etal-2023-enabling, khattab2023demonstratesearchpredict},  databases~\citep{pourreza2023dinsql, nan-etal-2023-enhancing, zhang2023reactable}, and graph databases~\citep{xu-etal-2023-fine, luo2023chatkbqa, li-etal-2023-shot}. 

In practice, we need to fully leverage hybrid data sources. For instance, Wikipedia alone offers a wealth of knowledge through nearly 7M free-text articles; many of these articles contain structured information in tables and infoboxes; Wikidata is a knowledge graph containing over 17 billion triples. This paper investigates how to leverage LLMs to answer questions on all the different types of data.
\begin{figure}[ht]
\centering
\includegraphics[width=0.45\textwidth]{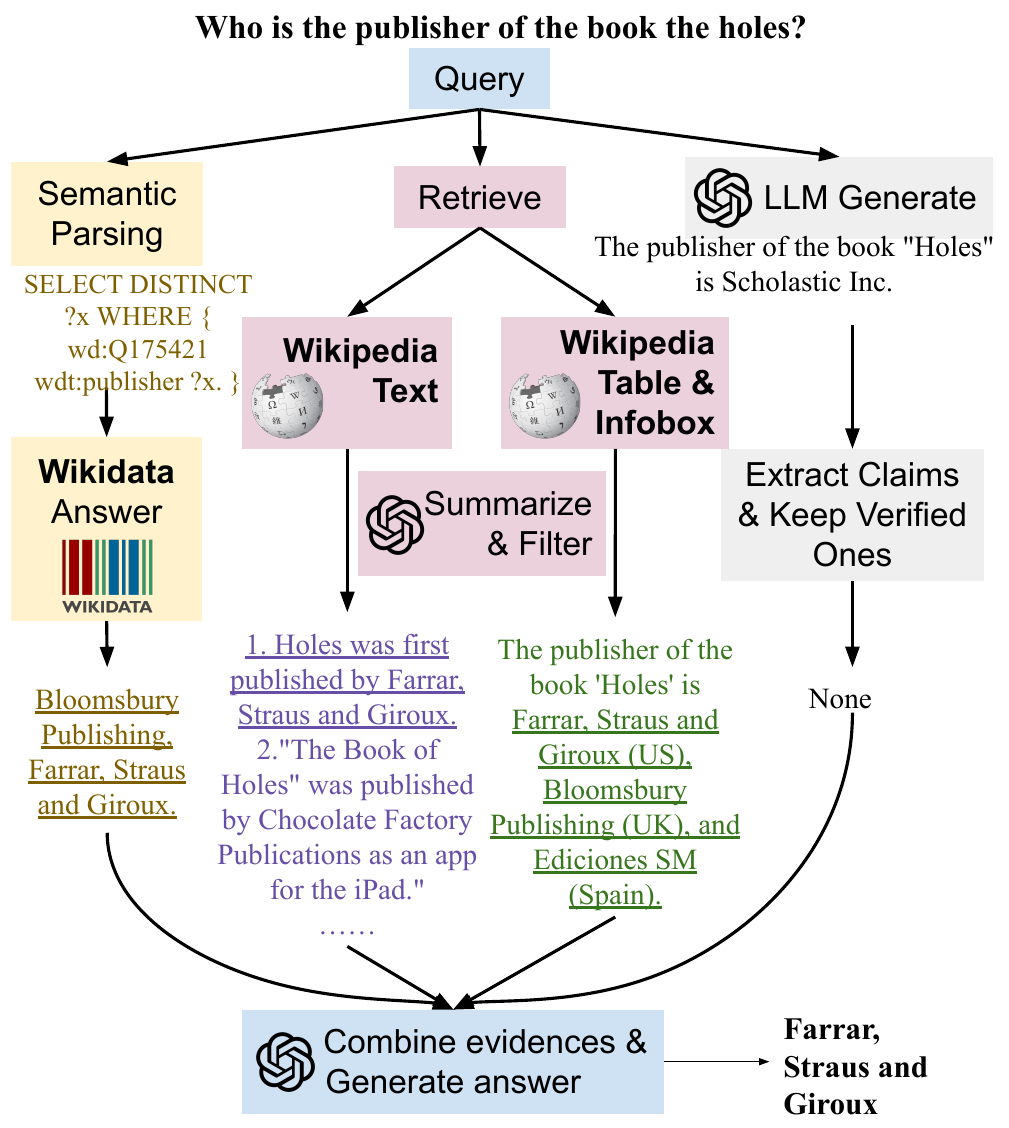}
\caption{Given an input query, \system gathers factual information from four sources to generate a prediction. In parallel, we parse the query to logical form to query Wikidata (left), run retrieval to find information from Wikipedia text, tables, and infoboxes (middle), and generate a response using LLM, only keeping a claim if it is verified (right). Finally, evidences are gathered to generate an answer.
\vspace{-0.3in}
}
\label{fig:overview}
\end{figure}


The premise of this paper is that we need hybrid data and we need hybrid access methods.
Our main contribution is twofold. First, we introduce a hybrid LLM-based system (Fig.~\ref{fig:overview}), \system, that combines information retrieval with semantic parsing for question answering and achieves SOTA of 56.5\% exact match rate on \compmix, the most comprehensive open-domain QA dataset on heterogeneous sources.
Second, we show via careful evaluation and analysis that measuring the accuracy of LLM-based QA systems with the exact-match metric against hand-annotated answers is obsolete. By using evaluation methods closer to human judgment, we find that \system is more than 90\% accurate on \compmix, surprisingly higher than traditional methods would suggest.

\section{Related Work}


TextQA, TableQA. and KBQA have all been individually studied extensively~\citep[\textit{inter alia}]{zhao2023survey, lu2024large, Pan_2024}. However, the task of answering questions from two or more sources, known as heterogeneous QA, is under-studied. Some literature investigate two of the three sources, including those on closed domain~\citep{miller-etal-2016-key, chen-etal-2020-hybridqa, DBLP:journals/corr/abs-2108-08614, liu2023suql, lei-etal-2023-s3hqa} and open domain~\citep{chen2021open, zhao2023divknowqa, han-gardent-2023-generating, ma-etal-2022-open-domain, ma-etal-2023-chain}, but very limited existing work experiments on all three.

\convmix~\citep{christmann2023conversational} collected the first conversational QA dataset that requires knowledge from all three heterogeneous sources. Crowdworkers were asked to pick an entity of their interest and find the answer from one of the Wiki sources - Wikidata, Wikipedia text, Wikipedia tables, or Wikipedia infoboxes.
\citet{christmann2023compmix} later collated the completed conversations to derive the \compmix dataset with 9410 self-contained question-answer pairs.


\citet{oguz-etal-2022-unik}, \citet{ma-etal-2022-open}, and \citet{christmann2023conversational} proposed pipelines to answer questions from all three sources, by linearizing all structured information and applying text retrieval methods.
\citet{christmann2023compmix}, on the other hand, unifies all the sources by representing all relevant information in a knowledge graph and uses GNN message passing to find the answer. The former gives up the advantage of using formal query languages on structured data, which can support operations such as ranking and averaging. The latter gives up the advantage of the expressiveness and versatility of free-text knowledge representation.
Concurrent with our work, \citet{Lehmann2024} adopts another view that breaks down QA solution processes as tool calls and thoughts. They propose a human-like approach that teaches LLMs to gather heterogeneous information by imitating how humans use retrieval tools, which requires human-annotated demonstrations.


\section{\system}

\system is a hybrid QA pipeline that takes advantage of both structured and unstructured information. We obtain evidence from heterogeneous sources in parallel, including structured knowledge bases, plain text,  linearized tables / infoboxes, and LLM-generated claims that are verified, and gather those evidence to generate the final answer using a few-shot LLM (Fig. \ref{fig:overview}).

\begin{figure}[ht]
\centering
\includegraphics[width=0.45\textwidth]{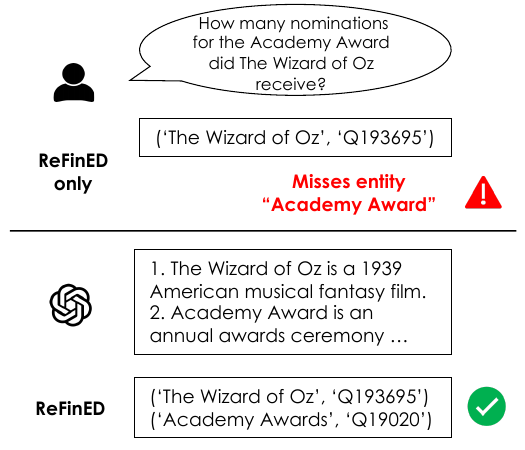}
\caption{An example with a failure case of ReFinED and our entity linking module correcting the failure.}
\label{fig:entity-linking}
\end{figure}
\subsection{Knowledge Base}
\label{sec:knowledge_base}
\citet{xu-etal-2023-fine} proposes a semantic parsing framework for Wikidata. By integrating a named entity linker and a fine-tuned LLaMA trained with modified SPARQL, they establish a strong baseline on the WikiWebQuestions dataset. We adopt their methodology as the interface to WikiData.

As noted by \citet{xu-etal-2023-fine}, most of the semantic parsing errors are due to the failure of the entity linker model ReFinED~\citep{ayoola-etal-2022-refined}. To improve on their approach, we propose a novel entity linking method where we first ask an LLM to detect entity mentions and generate a brief (maximum 10 words) description of each detected entity. We then feed the list of detected entities and descriptions to ReFinED to obtain the corresponding Wikidata entity IDs. Leveraging the world knowledge of an LLM in this fashion provides an additional mechanism to detect entity mentions and provide more context for ReFinED to disambiguate and link entities. Figure~\ref{fig:entity-linking} illustrates how our entity linking module works. Here, ReFinED alone fails to identify the entity ``Academy Award'', but it succeeds with LLM-provided context.



\subsection{Text}
Retrieval-augmented generation is a common approach for grounding LLMs in textual knowledge sources like Wikipedia. To avoid LLM hallucination, \citet{semnani-etal-2023-wikichat} proposes the \wikichat pipeline that combines retrieval with verification of LLM-generated response, achieving significantly higher factual accuracy than GPT-4. We adopt a similar approach when handling text.

We first extract Wikipedia text using Wikiextractor~\footnote{https://github.com/attardi/wikiextractor}. ColBERT~\citep{santhanam-etal-2022-colbertv2} is used to retrieve Wikipedia passages that may answer a given query, and each of the top-k retrieved passages goes through a few-shot LLM summarizer.

As shown in the rightmost path of Figure~\ref{fig:overview}, similar to WikiChat, \system also makes use of the internal factual knowledge of LLMs by first generating a response and then verifying the claims made in the response using retrieved information, retaining only grounded claims.


\subsection{Tables and Infoboxes}

Most NLP research using Wikipedia simply ignores the embedded tables and infoboxes, as extraction and preprocessing are challenging.
With the help of tools such as WikiTextParser~\footnote{https://github.com/5j9/wikitextparser} and regex matching, we programmatically extract 9 million tables and infoboxes from Wikipedia pages and linearize them so that they can be encoded as a set of ColBERT \citep{santhanam-etal-2022-colbertv2} index for retrieval. Being linearized, the retrieved item can then be read by LLMs directly.

Below is an example linearized table from the Wikipedia article ``Arundhati Roy'':
\begin{quote}
\small
    Fiction ; 
    \textbf{No.}: 1, 
    \textbf{Title}: ``The God of Small Things'', \textbf{Publisher}: Flamingo, \textbf{Year}: 1997, \textbf{ISBN}: ISBNT|0-00-655068-1.<tr> \textbf{No.}: 2, \textbf{Title}: ``The Ministry of Utmost Happiness'', 
    \textbf{Publisher}: Hamish Hamilton, 
    \textbf{Year}: 2017, 
    \textbf{ISBN}: ISBNT|0-241-30397-4 .<tr>
\end{quote}

For each table, we include the section title, two preceding sentences, and two succeeding sentences of the table as additional context, if there are any in the current section. Table rows are formatted as ``column\_name: cell\_content, ...'' with ``<tr>'' as the row separator.

Since ColBERT is pretrained with textual passages and not tables, we finetune ColBERT for table retrieval. After retrieval, the retrieved table is then fed into a few-shot LLM to extract information directly relevant to the query.





\subsection{Putting it Together}

At the final stage, we gather and combine evidence from all sources.
The answer from Wikidata is formatted as ``Wikidata says the answer to <query> is: <answer>.''
The retrieved text and tables/infoboxes each goes through an LLM summarization prompt, as mentioned earlier, attempting to extract relevant information from each retrieved item.
The verified claim(s) from the LLM-generated answer (if any) is also added to the evidence pool.

Finally, all evidence is fed to a few-shot LLM prompt to generate a single answer to the query. In some cases the answer may be contained in more than one information source, and such redundancy can help reduce errors introduced in earlier stages of the pipeline.

\begin{table*}[ht!]
    \small
    \centering
    \begin{tabular}{lccccccc}
        \toprule
         & \multicolumn{2}{c}{\textbf{Exact Match}} & \multicolumn{2}{c}{\textbf{Superset}} & \multicolumn{2}{c}{\textbf{GPT-4 Match}} & \textbf{Platinum}  \\
         \cmidrule(lr){2-3} \cmidrule(lr){4-5} \cmidrule(lr){6-7} \cmidrule(lr){8-8}
         & dev & test & dev & test & dev & test & dev (100)* \\
        \midrule
        CONVINSE \citep{christmann2023conversational} & -- & 40.7\% & -- & -- & -- & -- & --\\
        UniK-QA \citep{oguz-etal-2022-unik} & -- & 44.0\% & -- & -- & -- & -- & --\\
        EXPLAINGNN \citep{christmann2023explainable} & -- & 44.2\% & -- & -- & -- & -- & -- \\
        GPT-3 (text-davinci-003) & -- & 50.2\% & -- & -- & -- & -- & -- \\  
        GPT-3.5 (turbo-instruct) & 36.4\% & 36.1\% & 53.2\% & 54.2\% & 68.0\% & 69.9\% & 74\%\\
        GPT-4 & 53.0\% & 52.8\% & 60.9\% & 62.0\% & 76.7\% & 78.4\% & 81\%\\
        \midrule
        \system (LLaMA-7B) & 53.8\% & 51.7\% & 61.7\% & 60.5\% & 69.8\% & 70.4\% & 77\% \\
        \system (GPT-3.5) & \textbf{58.5\%} & 55.6\% & 67.7\% & 65.6\% & 76.9\% & 75.3\% & 84\%\\
        \system (GPT-4) & 57.3\% & \textbf{56.5\%} & \textbf{70.2\%} & \textbf{70.0\%} & \textbf{80.8\%} & \textbf{81.9\%} & \textbf{92\%} \\
        \bottomrule
    \end{tabular}
    \caption{Main results on the \compmix development and test set. UniK-QA and GPT-3 (text-davinci-003) results are from \citet{christmann2023compmix}. We use the same zero-shot generation prompt published by \citet{christmann2023compmix} to evaluate GPT-3.5 (turbo-instruct) and GPT-4. \\ *: Platinum results are obtained by an expert manually relabeling and evaluating the first 100 development set examples.}
        \vspace{-1em}
    \label{table:main-results}
\end{table*}

\begin{table}[ht!]
\small
\centering
\begin{tabular}{lccc}
\hline
 & \textbf{EM} & \textbf{Superset} & \textbf{GPT-4 Match} \\ \hline
Text        & 53.8\%            & 61.6\% & 71.1\% \\
Tables      & 48.9\%            & 59.5\% & 65.9\% \\
KB          & 32.9\%            & 40.4\% & -- \\
Text+Tables & 55.6\%            & 65.4\% &  74.5\%\\
Text+Tables+KB & \textbf{58.5\%}   & \textbf{67.7\%} & \textbf{76.9\%} \\ \hline
\end{tabular}
\caption{\system (GPT-3.5) ablation results on the \compmix development set, for using different knowledge sources. Results on ``KB'' are derived by directly comparing generated QID(s) against gold QID(s), while other methods are by string comparisons.}
\vspace{-1em}
\label{table:ablation}
\end{table}

\section{Experiments}

We evaluate \system on the \compmix development and test sets, which contain 1680 and 2764 questions respectively.

For querying Wikidata, we use the LLaMA-7B semantic parser from \citet{xu-etal-2023-fine} trained on both WikiWebQuestions and QALD-7~\citep{qald7}. We use GPT-3.5 as the LLM in our entity linking module.

We experiment with LLaMA-7B, GPT-3.5-turbo-instruct, and GPT-4\footnote{We access GPT models via the Microsoft Azure OpenAI API. We use the GPT-4 snapshot from June 13th, 2023.}, respectively, as the LLM backbone in all the stages for handling retrieved evidences and for answer generation. We use few-shot prompts for GPT-3.5 and GPT-4, and use the LLaMA model from \citet{semnani-etal-2023-wikichat}, which is distilled from the teacher GPT-4.

To fine-tune the ColBERT table retriever, we obtain training data from the NQ-Tables dataset~\cite{herzig-etal-2021-open}, where each example matches one gold table to a query. For each positive example, we sample 10 negative tables to obtain a total of 95K training triplets. We confirmed on the NQ-Tables dataset that the fine-tuned version improves table retrieval Recall@3 by 10\%.

\paragraph{Evaluation Metrics.}
\citet{bulian-etal-2022-tomayto} and \citet{kamalloo-etal-2023-evaluating} have established that exact match (EM) against gold answers, which is commonly used for evaluating QA systems, cannot evaluate generative models properly as they often generate lexically different, but semantically equivalent answers.  
To properly assess our approach, we introduce two additional evaluation metrics: (1) Superset: whether the gold answer is a substring of the generated answer, as the latter tends to spell out the answer in long form and may include a more complete answer. (2) GPT-4 Matching: using GPT-4 with a few-shot prompt to determine whether the generated answer matches the gold, similar to ~\citet{kamalloo-etal-2023-evaluating}.

Moreover, datasets may have ambiguous queries or even wrong annotations. To assess the quality of \compmix, we sample 100 questions and carefully use online information sources to find the answers and decide if the generated answers are correct. We refer to this metric as \emph{platinum} evaluation. 
GPT-4 for answer matching eliminates formatting issues, providing a result closer to human judgment than EM, while platinum evaluation further considers annotation issues and eliminates errors caused by ambiguous queries.

\section{Results}


\system (GPT-4) achieves 56.5\% EM rate on the test set of \compmix, improving on the previously reported state-of-the-art (GPT-3) by 6.3\%, and improves on the GPT-4 baseline by 3.7\% (Table \ref{table:main-results}). 
\system (GPT-3.5) also improves upon all of the baselines. We note that the EM scores of the GPT-3.5-turbo-instruct baseline are low because this model tends to be more verbose.


\system (GPT-4) achieves 81.9\% test set accuracy by GPT-4 matching, and 92\% \emph{platinum} accuracy on the 100 development set examples. Of the 8 errors cases, 3 have unanswerable questions (e.g. ``FC Cincinnati soccer club?''), thus the true accuracy rate is 92/97 (94\%). 

\paragraph{Ablations.}
We evaluate the contribution of each knowledge source by ablating different parts of the system (Table~\ref{table:ablation}). Using text alone already outperforms the previous SOTA, with each additional source further improving the result. Note that for many questions, information exists in multiple sources; the relatively little contribution from Wikidata and tables reflects mainly on the makeup of \compmix, not their value as knowledge sources. 
For detailed experimental results on our Wikidata entity linking approach, see Appendix \ref{sec:wikidata-experiments}.



\begin{figure}[ht]
\centering
\includegraphics[width=0.45\textwidth]{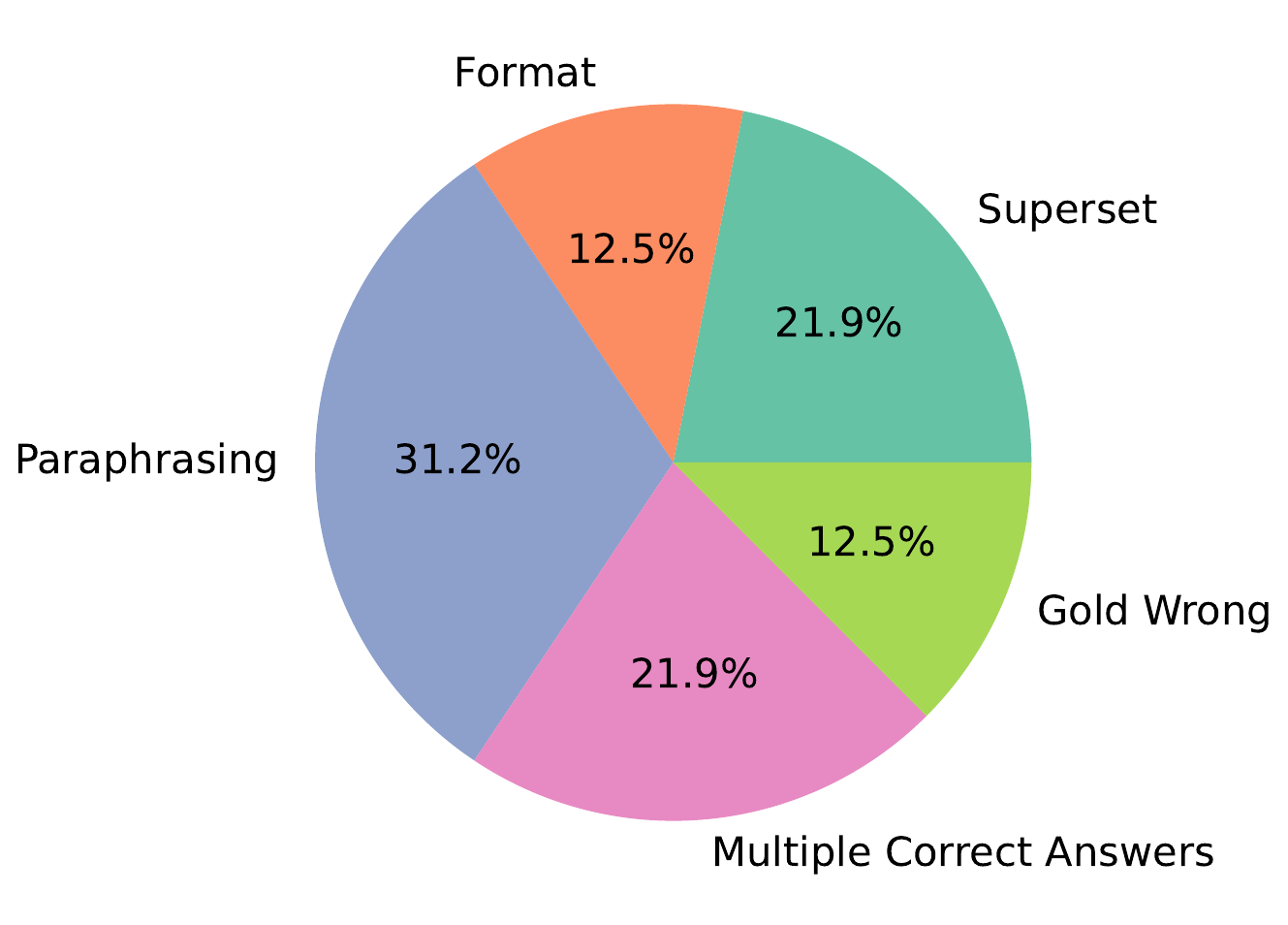}
\caption{Evaluation issues resolved within the gap between EM and \emph{platinum}.}
\label{fig:eval-pie-chart}
\end{figure}

\paragraph{Human Evaluation}
We examine how our human ``Platinum'' evaluation (92\%) differs from the EM metric (60\%) on our sample of 100 cases. 
We report findings on the \system (GPT-4) responses, specifically in 32 annotated examples where EM fails but the response is indeed correct (Figure~\ref{fig:eval-pie-chart}).
Out of the 32 discrepancies, the unsophisticated ``Superset'' metric resolves 7, and GPT-4 matching resolves an additional 14. 
Platinum evaluation identifies that 4 questions have incorrect gold labels, and 7 questions are ambiguous and the generated answers are correct though different from the gold. 
We include examples for each of these resolved cases in Appendix~\ref{sec:appendix-platinum-resolve}.


Of the 5 true errors, one is because \system cannot find the answer in any of the four information sources; 
in the other 4 cases, the answer generator cannot identify the correct answer retrieved due to conflicting or misleading evidence.
See Appendix~\ref{sec:error-analysis-appendix} for details on each error case.

\section{Conclusion}
We propose \system, a hybrid open-domain question-answering system that combines semantic parsing and information retrieval to handle structured and unstructured data.

\system achieves an exact match rate improvement of 6.3\% over the prior state-of-the-art on the \compmix dataset. More importantly, we show that our approach is likely to reach an accuracy of over 90\%, if we account for differences in the answer wording and incompleteness/errors in gold labels. 
This, however, does not mean open-domain QA is solved. Further research is needed to handle open-domain questions that require complex structured queries or composition of answers from multiple information sources, none of which are included in \compmix. 




\section*{Limitations}

This work focuses specifically on open-domain QA with heterogeneous knowledge sources, and we only report results on the \compmix dataset due to the limited availability of high-quality datasets in this domain.
A natural future work is to develop more diverse and advanced datasets that further push the need to utilize each knowledge source.

We evaluate on single-turn QA and do not work with conversations in this paper, and \system can be extended to handle fact-based conversational questions or even chitchat that involves facts.

We have a relatively small sample size for human evaluation, because the expert manually checks the correctness of each example with Internet searches, which is labor-intensive. However, we acknowledge that a larger sample size would increase the statistical confidence of our evaluation.

Finally, we note that a number of Wikipedia tables are not well-formatted after preprocessing and linearization. Since Wikipedia tables are embedded as HTML elements that allow for idiosyncrasies like a table with one cell spanning multiple columns or color-highlighted cells, some are hard to parse correctly. Solving such edge cases engineering-wise would further improve TableQA.


\section*{Ethical Considerations}
To facilitate reproducibility and continued research, we will make the code available upon publication.

No new datasets were gathered specifically for this study, and we did not employ crowd-sourced labor. 
We use Wikipedia data under the terms of the Creative Commons Attribution-ShareAlike 4.0 International License (CC BY-SA) and the GNU Free Documentation License (GFDL). Wikidata is under Creative Commons CC0 License, which is equivalent to public domain. The \compmix benchmark is licensed under a Creative Commons Attribution 4.0 International License. We use the benchmark as it is intended.

The experimental phase involved approximately 80 hours of computation time on an NVIDIA A100 GPU to fine-tune the retrieval model and index Wikipedia content. We reused the LLaMA-7B model trained in prior work, thus avoiding extra GPU usage.

We do not anticipate adverse effects stemming from the proposed methods in this study.

\bibliography{anthology,custom}

\begin{thebibliography}{42}
\expandafter\ifx\csname natexlab\endcsname\relax\def\natexlab#1{#1}\fi

\bibitem[{Asai et~al.(2022)Asai, Gardner, and Hajishirzi}]{asai-etal-2022-evidentiality}
Akari Asai, Matt Gardner, and Hannaneh Hajishirzi. 2022.
\newblock \href {https://doi.org/10.18653/v1/2022.naacl-main.162} {Evidentiality-guided generation for knowledge-intensive {NLP} tasks}.
\newblock In \emph{Proceedings of the 2022 Conference of the North American Chapter of the Association for Computational Linguistics: Human Language Technologies}, pages 2226--2243, Seattle, United States. Association for Computational Linguistics.

\bibitem[{Asai et~al.(2020)Asai, Hashimoto, Hajishirzi, Socher, and Xiong}]{Asai2020Learning}
Akari Asai, Kazuma Hashimoto, Hannaneh Hajishirzi, Richard Socher, and Caiming Xiong. 2020.
\newblock \href {https://openreview.net/forum?id=SJgVHkrYDH} {Learning to retrieve reasoning paths over wikipedia graph for question answering}.
\newblock In \emph{International Conference on Learning Representations}.

\bibitem[{Ayoola et~al.(2022)Ayoola, Tyagi, Fisher, Christodoulopoulos, and Pierleoni}]{ayoola-etal-2022-refined}
Tom Ayoola, Shubhi Tyagi, Joseph Fisher, Christos Christodoulopoulos, and Andrea Pierleoni. 2022.
\newblock \href {https://doi.org/10.18653/v1/2022.naacl-industry.24} {{R}e{F}in{ED}: An efficient zero-shot-capable approach to end-to-end entity linking}.
\newblock In \emph{Proceedings of the 2022 Conference of the North American Chapter of the Association for Computational Linguistics: Human Language Technologies: Industry Track}, pages 209--220, Hybrid: Seattle, Washington + Online. Association for Computational Linguistics.

\bibitem[{Bulian et~al.(2022)Bulian, Buck, Gajewski, B{\"o}rschinger, and Schuster}]{bulian-etal-2022-tomayto}
Jannis Bulian, Christian Buck, Wojciech Gajewski, Benjamin B{\"o}rschinger, and Tal Schuster. 2022.
\newblock \href {https://doi.org/10.18653/v1/2022.emnlp-main.20} {Tomayto, tomahto. beyond token-level answer equivalence for question answering evaluation}.
\newblock In \emph{Proceedings of the 2022 Conference on Empirical Methods in Natural Language Processing}, pages 291--305, Abu Dhabi, United Arab Emirates. Association for Computational Linguistics.

\bibitem[{Chen et~al.(2017)Chen, Fisch, Weston, and Bordes}]{chen-etal-2017-reading}
Danqi Chen, Adam Fisch, Jason Weston, and Antoine Bordes. 2017.
\newblock \href {https://doi.org/10.18653/v1/P17-1171} {Reading {W}ikipedia to answer open-domain questions}.
\newblock In \emph{Proceedings of the 55th Annual Meeting of the Association for Computational Linguistics (Volume 1: Long Papers)}, pages 1870--1879, Vancouver, Canada. Association for Computational Linguistics.

\bibitem[{Chen et~al.(2021)Chen, Chang, Schlinger, Wang, and Cohen}]{chen2021open}
Wenhu Chen, Ming-Wei Chang, Eva Schlinger, William~Yang Wang, and William~W. Cohen. 2021.
\newblock \href {https://openreview.net/forum?id=MmCRswl1UYl} {Open question answering over tables and text}.
\newblock In \emph{International Conference on Learning Representations}.

\bibitem[{Chen et~al.(2020)Chen, Zha, Chen, Xiong, Wang, and Wang}]{chen-etal-2020-hybridqa}
Wenhu Chen, Hanwen Zha, Zhiyu Chen, Wenhan Xiong, Hong Wang, and William~Yang Wang. 2020.
\newblock \href {https://doi.org/10.18653/v1/2020.findings-emnlp.91} {{H}ybrid{QA}: A dataset of multi-hop question answering over tabular and textual data}.
\newblock In \emph{Findings of the Association for Computational Linguistics: EMNLP 2020}, pages 1026--1036, Online. Association for Computational Linguistics.

\bibitem[{Christmann et~al.(2023{\natexlab{a}})Christmann, Roy, and Weikum}]{christmann2023compmix}
Philipp Christmann, Rishiraj~Saha Roy, and Gerhard Weikum. 2023{\natexlab{a}}.
\newblock \href {http://arxiv.org/abs/2306.12235} {Compmix: A benchmark for heterogeneous question answering}.

\bibitem[{Christmann et~al.(2022)Christmann, Saha~Roy, and Weikum}]{christmann2023conversational}
Philipp Christmann, Rishiraj Saha~Roy, and Gerhard Weikum. 2022.
\newblock \href {https://doi.org/10.1145/3477495.3531815} {Conversational question answering on heterogeneous sources}.
\newblock In \emph{Proceedings of the 45th International ACM SIGIR Conference on Research and Development in Information Retrieval}, SIGIR '22, page 144–154, New York, NY, USA. Association for Computing Machinery.

\bibitem[{Christmann et~al.(2023{\natexlab{b}})Christmann, Saha~Roy, and Weikum}]{christmann2023explainable}
Philipp Christmann, Rishiraj Saha~Roy, and Gerhard Weikum. 2023{\natexlab{b}}.
\newblock \href {https://doi.org/10.1145/3539618.3591682} {Explainable conversational question answering over heterogeneous sources via iterative graph neural networks}.
\newblock In \emph{Proceedings of the 46th International ACM SIGIR Conference on Research and Development in Information Retrieval}, SIGIR '23, page 643–653, New York, NY, USA. Association for Computing Machinery.

\bibitem[{Gao et~al.(2023)Gao, Yen, Yu, and Chen}]{gao-etal-2023-enabling}
Tianyu Gao, Howard Yen, Jiatong Yu, and Danqi Chen. 2023.
\newblock \href {https://doi.org/10.18653/v1/2023.emnlp-main.398} {Enabling large language models to generate text with citations}.
\newblock In \emph{Proceedings of the 2023 Conference on Empirical Methods in Natural Language Processing}, pages 6465--6488, Singapore. Association for Computational Linguistics.

\bibitem[{Han and Gardent(2023)}]{han-gardent-2023-generating}
Kelvin Han and Claire Gardent. 2023.
\newblock \href {https://aclanthology.org/2023.ijcnlp-main.19} {Generating and answering simple and complex questions from text and from knowledge graphs}.
\newblock In \emph{Proceedings of the 13th International Joint Conference on Natural Language Processing and the 3rd Conference of the Asia-Pacific Chapter of the Association for Computational Linguistics (Volume 1: Long Papers)}, pages 285--304, Nusa Dua, Bali. Association for Computational Linguistics.

\bibitem[{Herzig et~al.(2021)Herzig, M{\"u}ller, Krichene, and Eisenschlos}]{herzig-etal-2021-open}
Jonathan Herzig, Thomas M{\"u}ller, Syrine Krichene, and Julian Eisenschlos. 2021.
\newblock \href {https://doi.org/10.18653/v1/2021.naacl-main.43} {Open domain question answering over tables via dense retrieval}.
\newblock In \emph{Proceedings of the 2021 Conference of the North American Chapter of the Association for Computational Linguistics: Human Language Technologies}, pages 512--519, Online. Association for Computational Linguistics.

\bibitem[{Izacard and Grave(2021)}]{izacard-grave-2021-leveraging}
Gautier Izacard and Edouard Grave. 2021.
\newblock \href {https://doi.org/10.18653/v1/2021.eacl-main.74} {Leveraging passage retrieval with generative models for open domain question answering}.
\newblock In \emph{Proceedings of the 16th Conference of the European Chapter of the Association for Computational Linguistics: Main Volume}, pages 874--880, Online. Association for Computational Linguistics.

\bibitem[{Jiang et~al.(2023)Jiang, Xu, Gao, Sun, Liu, Dwivedi-Yu, Yang, Callan, and Neubig}]{jiang-etal-2023-active}
Zhengbao Jiang, Frank Xu, Luyu Gao, Zhiqing Sun, Qian Liu, Jane Dwivedi-Yu, Yiming Yang, Jamie Callan, and Graham Neubig. 2023.
\newblock \href {https://doi.org/10.18653/v1/2023.emnlp-main.495} {Active retrieval augmented generation}.
\newblock In \emph{Proceedings of the 2023 Conference on Empirical Methods in Natural Language Processing}, pages 7969--7992, Singapore. Association for Computational Linguistics.

\bibitem[{Kamalloo et~al.(2023)Kamalloo, Dziri, Clarke, and Rafiei}]{kamalloo-etal-2023-evaluating}
Ehsan Kamalloo, Nouha Dziri, Charles Clarke, and Davood Rafiei. 2023.
\newblock \href {https://doi.org/10.18653/v1/2023.acl-long.307} {Evaluating open-domain question answering in the era of large language models}.
\newblock In \emph{Proceedings of the 61st Annual Meeting of the Association for Computational Linguistics (Volume 1: Long Papers)}, pages 5591--5606, Toronto, Canada. Association for Computational Linguistics.

\bibitem[{Khattab et~al.(2021)Khattab, Potts, and Zaharia}]{khattab-etal-2021-relevance}
Omar Khattab, Christopher Potts, and Matei Zaharia. 2021.
\newblock \href {https://doi.org/10.1162/tacl_a_00405} {Relevance-guided supervision for {O}pen{QA} with {C}ol{BERT}}.
\newblock \emph{Transactions of the Association for Computational Linguistics}, 9:929--944.

\bibitem[{Khattab et~al.(2023)Khattab, Santhanam, Li, Hall, Liang, Potts, and Zaharia}]{khattab2023demonstratesearchpredict}
Omar Khattab, Keshav Santhanam, Xiang~Lisa Li, David Hall, Percy Liang, Christopher Potts, and Matei Zaharia. 2023.
\newblock \href {http://arxiv.org/abs/2212.14024} {Demonstrate-search-predict: Composing retrieval and language models for knowledge-intensive nlp}.

\bibitem[{Lee et~al.(2019)Lee, Chang, and Toutanova}]{lee-etal-2019-latent}
Kenton Lee, Ming-Wei Chang, and Kristina Toutanova. 2019.
\newblock \href {https://doi.org/10.18653/v1/P19-1612} {Latent retrieval for weakly supervised open domain question answering}.
\newblock In \emph{Proceedings of the 57th Annual Meeting of the Association for Computational Linguistics}, pages 6086--6096, Florence, Italy. Association for Computational Linguistics.

\bibitem[{Lehmann et~al.(2024)Lehmann, Bhandiwad, Gattogi, and Vahdati}]{Lehmann2024}
Jens Lehmann, Dhananjay Bhandiwad, Preetam Gattogi, and Sahar Vahdati. 2024.
\newblock \href {https://www.amazon.science/publications/beyond-boundaries-a-human-like-approach-for-question-answering-over-structured-and-unstructured-information-sources} {Beyond boundaries: A human-like approach for question answering over structured and unstructured information sources}.
\newblock \emph{Transactions of the Association for Computational Linguistics (TACL)}.

\bibitem[{Lei et~al.(2023)Lei, Li, Wei, He, Huang, Zhao, and Liu}]{lei-etal-2023-s3hqa}
Fangyu Lei, Xiang Li, Yifan Wei, Shizhu He, Yiming Huang, Jun Zhao, and Kang Liu. 2023.
\newblock \href {https://doi.org/10.18653/v1/2023.acl-short.147} {{S}3{HQA}: A three-stage approach for multi-hop text-table hybrid question answering}.
\newblock In \emph{Proceedings of the 61st Annual Meeting of the Association for Computational Linguistics (Volume 2: Short Papers)}, pages 1731--1740, Toronto, Canada. Association for Computational Linguistics.

\bibitem[{Li et~al.(2023)Li, Ma, Zhuang, Gu, Su, and Chen}]{li-etal-2023-shot}
Tianle Li, Xueguang Ma, Alex Zhuang, Yu~Gu, Yu~Su, and Wenhu Chen. 2023.
\newblock \href {https://doi.org/10.18653/v1/2023.acl-long.385} {Few-shot in-context learning on knowledge base question answering}.
\newblock In \emph{Proceedings of the 61st Annual Meeting of the Association for Computational Linguistics (Volume 1: Long Papers)}, pages 6966--6980, Toronto, Canada. Association for Computational Linguistics.

\bibitem[{Liu et~al.(2023)Liu, Xu, Tjangnaka, Semnani, Yu, Dávid, and Lam}]{liu2023suql}
Shicheng Liu, Jialiang Xu, Wesley Tjangnaka, Sina~J. Semnani, Chen~Jie Yu, Gui Dávid, and Monica~S. Lam. 2023.
\newblock \href {http://arxiv.org/abs/2311.09818} {{SUQL}: Conversational search over structured and unstructured data with large language models}.

\bibitem[{Lu et~al.(2024)Lu, Zhang, Zhang, and Chen}]{lu2024large}
Weizheng Lu, Jiaming Zhang, Jing Zhang, and Yueguo Chen. 2024.
\newblock \href {http://arxiv.org/abs/2402.05121} {Large language model for table processing: A survey}.

\bibitem[{Luo et~al.(2023)Luo, E, Tang, Peng, Guo, Zhang, Ma, Dong, Song, and Lin}]{luo2023chatkbqa}
Haoran Luo, Haihong E, Zichen Tang, Shiyao Peng, Yikai Guo, Wentai Zhang, Chenghao Ma, Guanting Dong, Meina Song, and Wei Lin. 2023.
\newblock \href {http://arxiv.org/abs/2310.08975} {Chatkbqa: A generate-then-retrieve framework for knowledge base question answering with fine-tuned large language models}.

\bibitem[{Ma et~al.(2022{\natexlab{a}})Ma, Cheng, Liu, Nyberg, and Gao}]{ma-etal-2022-open-domain}
Kaixin Ma, Hao Cheng, Xiaodong Liu, Eric Nyberg, and Jianfeng Gao. 2022{\natexlab{a}}.
\newblock \href {https://doi.org/10.18653/v1/2022.findings-emnlp.392} {Open-domain question answering via chain of reasoning over heterogeneous knowledge}.
\newblock In \emph{Findings of the Association for Computational Linguistics: EMNLP 2022}, pages 5360--5374, Abu Dhabi, United Arab Emirates. Association for Computational Linguistics.

\bibitem[{Ma et~al.(2022{\natexlab{b}})Ma, Cheng, Liu, Nyberg, and Gao}]{ma-etal-2022-open}
Kaixin Ma, Hao Cheng, Xiaodong Liu, Eric Nyberg, and Jianfeng Gao. 2022{\natexlab{b}}.
\newblock \href {https://doi.org/10.18653/v1/2022.acl-long.113} {Open domain question answering with a unified knowledge interface}.
\newblock In \emph{Proceedings of the 60th Annual Meeting of the Association for Computational Linguistics (Volume 1: Long Papers)}, pages 1605--1620, Dublin, Ireland. Association for Computational Linguistics.

\bibitem[{Ma et~al.(2023)Ma, Cheng, Zhang, Liu, Nyberg, and Gao}]{ma-etal-2023-chain}
Kaixin Ma, Hao Cheng, Yu~Zhang, Xiaodong Liu, Eric Nyberg, and Jianfeng Gao. 2023.
\newblock \href {https://doi.org/10.18653/v1/2023.acl-long.89} {Chain-of-skills: A configurable model for open-domain question answering}.
\newblock In \emph{Proceedings of the 61st Annual Meeting of the Association for Computational Linguistics (Volume 1: Long Papers)}, pages 1599--1618, Toronto, Canada. Association for Computational Linguistics.

\bibitem[{Miller et~al.(2016)Miller, Fisch, Dodge, Karimi, Bordes, and Weston}]{miller-etal-2016-key}
Alexander Miller, Adam Fisch, Jesse Dodge, Amir-Hossein Karimi, Antoine Bordes, and Jason Weston. 2016.
\newblock \href {https://doi.org/10.18653/v1/D16-1147} {Key-value memory networks for directly reading documents}.
\newblock In \emph{Proceedings of the 2016 Conference on Empirical Methods in Natural Language Processing}, pages 1400--1409, Austin, Texas. Association for Computational Linguistics.

\bibitem[{Nan et~al.(2023)Nan, Zhao, Zou, Ri, Tae, Zhang, Cohan, and Radev}]{nan-etal-2023-enhancing}
Linyong Nan, Yilun Zhao, Weijin Zou, Narutatsu Ri, Jaesung Tae, Ellen Zhang, Arman Cohan, and Dragomir Radev. 2023.
\newblock \href {https://doi.org/10.18653/v1/2023.findings-emnlp.996} {Enhancing text-to-{SQL} capabilities of large language models: A study on prompt design strategies}.
\newblock In \emph{Findings of the Association for Computational Linguistics: EMNLP 2023}, pages 14935--14956, Singapore. Association for Computational Linguistics.

\bibitem[{Oguz et~al.(2022)Oguz, Chen, Karpukhin, Peshterliev, Okhonko, Schlichtkrull, Gupta, Mehdad, and Yih}]{oguz-etal-2022-unik}
Barlas Oguz, Xilun Chen, Vladimir Karpukhin, Stan Peshterliev, Dmytro Okhonko, Michael Schlichtkrull, Sonal Gupta, Yashar Mehdad, and Scott Yih. 2022.
\newblock \href {https://doi.org/10.18653/v1/2022.findings-naacl.115} {{U}ni{K}-{QA}: Unified representations of structured and unstructured knowledge for open-domain question answering}.
\newblock In \emph{Findings of the Association for Computational Linguistics: NAACL 2022}, pages 1535--1546, Seattle, United States. Association for Computational Linguistics.

\bibitem[{Pan et~al.(2024)Pan, Luo, Wang, Chen, Wang, and Wu}]{Pan_2024}
Shirui Pan, Linhao Luo, Yufei Wang, Chen Chen, Jiapu Wang, and Xindong Wu. 2024.
\newblock \href {https://doi.org/10.1109/tkde.2024.3352100} {Unifying large language models and knowledge graphs: A roadmap}.
\newblock \emph{IEEE Transactions on Knowledge and Data Engineering}, page 1–20.

\bibitem[{Pourreza and Rafiei(2023)}]{pourreza2023dinsql}
Mohammadreza Pourreza and Davood Rafiei. 2023.
\newblock \href {http://arxiv.org/abs/2304.11015} {Din-sql: Decomposed in-context learning of text-to-sql with self-correction}.

\bibitem[{Pramanik et~al.(2021)Pramanik, Alabi, Roy, and Weikum}]{DBLP:journals/corr/abs-2108-08614}
Soumajit Pramanik, Jesujoba Alabi, Rishiraj~Saha Roy, and Gerhard Weikum. 2021.
\newblock \href {http://arxiv.org/abs/2108.08614} {{UNIQORN:} unified question answering over {RDF} knowledge graphs and natural language text}.
\newblock \emph{CoRR}, abs/2108.08614.

\bibitem[{Santhanam et~al.(2022)Santhanam, Khattab, Saad-Falcon, Potts, and Zaharia}]{santhanam-etal-2022-colbertv2}
Keshav Santhanam, Omar Khattab, Jon Saad-Falcon, Christopher Potts, and Matei Zaharia. 2022.
\newblock \href {https://doi.org/10.18653/v1/2022.naacl-main.272} {{C}ol{BERT}v2: Effective and efficient retrieval via lightweight late interaction}.
\newblock In \emph{Proceedings of the 2022 Conference of the North American Chapter of the Association for Computational Linguistics: Human Language Technologies}, pages 3715--3734, Seattle, United States. Association for Computational Linguistics.

\bibitem[{Semnani et~al.(2023)Semnani, Yao, Zhang, and Lam}]{semnani-etal-2023-wikichat}
Sina Semnani, Violet Yao, Heidi Zhang, and Monica Lam. 2023.
\newblock \href {https://doi.org/10.18653/v1/2023.findings-emnlp.157} {{W}iki{C}hat: Stopping the hallucination of large language model chatbots by few-shot grounding on {W}ikipedia}.
\newblock In \emph{Findings of the Association for Computational Linguistics: EMNLP 2023}, pages 2387--2413, Singapore. Association for Computational Linguistics.

\bibitem[{Usbeck et~al.(2017)Usbeck, Ngomo, Haarmann, Krithara, R{\"o}der, and Napolitano}]{qald7}
Ricardo Usbeck, Axel-Cyrille~Ngonga Ngomo, Bastian Haarmann, Anastasia Krithara, Michael R{\"o}der, and Giulio Napolitano. 2017.
\newblock 7th open challenge on question answering over linked data (qald-7).
\newblock In \emph{Semantic web evaluation challenge}, pages 59--69. Springer.

\bibitem[{Wang et~al.(2018)Wang, Yu, Guo, Wang, Klinger, Zhang, Chang, Tesauro, Zhou, and Jiang}]{Wang_Yu_Guo_Wang_Klinger_Zhang_Chang_Tesauro_Zhou_Jiang_2018}
Shuohang Wang, Mo~Yu, Xiaoxiao Guo, Zhiguo Wang, Tim Klinger, Wei Zhang, Shiyu Chang, Gerry Tesauro, Bowen Zhou, and Jing Jiang. 2018.
\newblock \href {https://doi.org/10.1609/aaai.v32i1.12053} {R\textasciicircum 3: Reinforced ranker-reader for open-domain question answering}.
\newblock \emph{Proceedings of the AAAI Conference on Artificial Intelligence}, 32(1).

\bibitem[{Xu et~al.(2023)Xu, Liu, Culhane, Pertseva, Wu, Semnani, and Lam}]{xu-etal-2023-fine}
Silei Xu, Shicheng Liu, Theo Culhane, Elizaveta Pertseva, Meng-Hsi Wu, Sina Semnani, and Monica Lam. 2023.
\newblock \href {https://doi.org/10.18653/v1/2023.emnlp-main.353} {Fine-tuned {LLM}s know more, hallucinate less with few-shot sequence-to-sequence semantic parsing over {W}ikidata}.
\newblock In \emph{Proceedings of the 2023 Conference on Empirical Methods in Natural Language Processing}, pages 5778--5791, Singapore. Association for Computational Linguistics.

\bibitem[{Zhang et~al.(2023)Zhang, Henkel, Floratou, Cahoon, Deep, and Patel}]{zhang2023reactable}
Yunjia Zhang, Jordan Henkel, Avrilia Floratou, Joyce Cahoon, Shaleen Deep, and Jignesh~M. Patel. 2023.
\newblock \href {http://arxiv.org/abs/2310.00815} {Reactable: Enhancing react for table question answering}.

\bibitem[{Zhao et~al.(2023{\natexlab{a}})Zhao, Zhou, Li, Tang, Wang, Hou, Min, Zhang, Zhang, Dong, Du, Yang, Chen, Chen, Jiang, Ren, Li, Tang, Liu, Liu, Nie, and Wen}]{zhao2023survey}
Wayne~Xin Zhao, Kun Zhou, Junyi Li, Tianyi Tang, Xiaolei Wang, Yupeng Hou, Yingqian Min, Beichen Zhang, Junjie Zhang, Zican Dong, Yifan Du, Chen Yang, Yushuo Chen, Zhipeng Chen, Jinhao Jiang, Ruiyang Ren, Yifan Li, Xinyu Tang, Zikang Liu, Peiyu Liu, Jian-Yun Nie, and Ji-Rong Wen. 2023{\natexlab{a}}.
\newblock \href {http://arxiv.org/abs/2303.18223} {A survey of large language models}.

\bibitem[{Zhao et~al.(2023{\natexlab{b}})Zhao, Liu, Niu, Wan, Yu, Joty, Zhou, and Yavuz}]{zhao2023divknowqa}
Wenting Zhao, Ye~Liu, Tong Niu, Yao Wan, Philip~S. Yu, Shafiq Joty, Yingbo Zhou, and Semih Yavuz. 2023{\natexlab{b}}.
\newblock \href {http://arxiv.org/abs/2310.20170} {Divknowqa: Assessing the reasoning ability of llms via open-domain question answering over knowledge base and text}.

\end{thebibliography}

\appendix



\clearpage



\section{Wikidata Experiments}
\label{sec:wikidata-experiments}


\citet{xu-etal-2023-fine} fine-tuned two LLaMAs on Wikidata. The training data for the first model consists solely of WikiWebQuestions~\citep{xu-etal-2023-fine}, while the other consists of the combination of WikiWebQuestions~\citep{xu-etal-2023-fine} and QALD-7~\citep{qald7}. We experiment with both models on the development set of \compmix, each with (1) entities predicted by ReFinED, (2) our entity linking approach with GPT-3.5 as the LLM (prompt in Figure \ref{tab:ned-prompt}), and (3) the dataset-provided oracle entities.

\begin{table*}[ht]
    \small
    \centering
    \setlength\tabcolsep{4pt}
    \begin{tabular}{lcc|cc}
        \toprule
         & \multicolumn{2}{c}{\textbf{Dev}} & \multicolumn{2}{c}{\textbf{Dev (KB subset)}} \\

         & EM & Superset & EM & Superset  \\
        \midrule
        WikiWebQuestions Only & & & & \\
        w/ ReFinED only entities & 29.1 & 36.5 & 39.8  & 46.0  \\
        w/ ReFinED $+$ GPT-3.5 entities & 31.3 & 38.8 & 43.8 & 50.4 \\
        w/ oracle entities & 33.9 & 42.3 & 46.2 & 52.8 \\
        \midrule
        WikiWebQuestions $+$ Qald-7 & & & & \\
        w/ ReFinED only entities & 29.5 & 36.6 & 41.0 & 47.6  \\
        w/ ReFinED $+$ GPT-3.5 entities & 32.9 & 40.4 & 46.8 & 53.0 \\
        w/ oracle entities & 35.5 & 43.1 & 49.0 & 55.4 \\
        \bottomrule
    \end{tabular}
    \caption{Wikidata semantic parsing experiment results on the \compmix development set. Comparison is made using entity IDs. Superset measures whether the model's predicted entities is a superset of the gold entities. \textbf{Dev (KB subset)} refers to the subset of the dataset where the annotators located the annotated answer from Wikidata.}
        \vspace{-1em}
    \label{tab:wikidata-ablation}
\end{table*}

As shown in Table \ref{tab:wikidata-ablation}, the model using entities predicted by our approach outperforms the model using the baseline ReFinED entities. It achieves considerably closer performance with the model using oracle entities. We also observed that the the model trained on both WikiWebQuestions and QALD-7 outperforms the model trained on WikiWebQuestions only.

\section{Details on Platinum Evaluation}
\label{sec:appendix-platinum-resolve}
Figure~\ref{fig:eval-pie-chart} shows the distribution of cases that we resolve using more advanced evaluation metrics. Numbers are reported on the first 100 dev examples with \system (GPT-4).




Examples of each evaluation error type can be found at Figure~\ref{lst:em_cannot_handle_1}, Figure~\ref{lst:em_cannot_handle_2}, Figure~\ref{lst:em_cannot_handle_3}, Figure~\ref{lst:em_cannot_handle_4}, and Figure~\ref{lst:em_cannot_handle_5}.

\section{Error Analysis}
\label{sec:error-analysis-appendix}

We include the five error cases after platinum evaluation in Figure~\ref{lst:platinum_wrong_1}, Figure~\ref{lst:platinum_wrong_2}, Figure~\ref{lst:platinum_wrong_3}, Figure~\ref{lst:platinum_wrong_4}, and Figure~\ref{lst:platinum_wrong_5}.


\subsection{Conflicting or Misleading Evidence}

\label{sec:conflicting_evidence}


We analyze the 388 error cases from \system(GPT-3.5) as determined by GPT-4 Matching. We separate out evidence retrieval errors from answer generation errors by identifying how often the gold answer appears in the evidences using a substring matching heuristic (Table~\ref{tab:gold_in_evidence}).

\begin{table}[ht]
\centering
\resizebox{0.65\columnwidth}{!}{%
\begin{tabular}{@{}lc@{}}
\toprule
\multicolumn{1}{c}{} & \textbf{\# Error Cases} \\ \midrule
All Error Cases      & 388 (100\%)       \\
Gold in Evidence     & 154 (39.69\%)        \\
Gold in KB           & 51 (13.14\%)         \\
Gold in Text         & 87 (22.42\%)         \\
Gold in Tables        & 72 (18.56\%)         \\ \bottomrule
\end{tabular}%
}
\caption{Numbers of error cases by category. The notation ``Gold~in~[source]'' stands for the gold answer existing as a substring in the particular [source].}
\label{tab:gold_in_evidence}
\end{table}

In 154 out of all 388 error cases, the system does not produce the gold answer despite the successful retrieval of evidence containing it. This observation indicates that a significant portion of the error cases are due to conflicting or misleading information in the evidence, where further improvements in selecting and merging evidences would be helpful. In the majority of the error cases (234 out of 388) where gold is not in the evidence, the system has no high-quality candidates to select from. Note, however, that this is an overestimate, due to the use of substring matching for deciding whether an evidence is correct or not.

In the breakdown of gold answer sources, the source that contains the most gold answers is Text (87 out of 154 cases), and Wikidata contains the least gold answers (51 out of 154 cases).





\subsection{Combiner Hallucination}
\label{sec:refinement_hallucination}
We investigate the ratio of generated answers that were hallucinated by our model. We manually checked the first 300 cases in our evaluation set and found 2 cases (0.67\%) where the model ignored the evidence and hallucinated an incorrect answer. This low ratio of hallucination highlights the faithfulness of our system to the evidence retrieved. We include these cases in Figure~\ref{lst:refinement_hallucination_1} and Figure~\ref{lst:refinement_hallucination_2}.

\begin{figure*}[ht]
\begin{lstlisting}[language=json]
{
    "idx": 64,
    "correct_hallucination": true,
    "question": "Nirvana was founded by who?",
    "gold": "Kurt Cobain",
    "answer_generated": "Kurt Cobain, Krist Novoselic, and Dave Grohl",
    "gold_sources": [
        "TEXT"
    ],
    "pred_sources": [],
    "evidences": [
        [
            "KB",
            "Wikidata says the answer to \"Nirvana was founded by who?\" is: ."
        ],
        [
            "TEXT",
            "Tan Sri Kong Hon Kong is the founder of Nirvana Asia Group, the largest integrated funeral service provider in Malaysia."
        ],
        [
            "TEXT",
            "Nirvana was founded by lead singer and guitarist Kurt Cobain and bassist Krist Novoselic in 1987."
        ],
        [
            "TEXT",
            "The founder of Buddhism, the Buddha, is believed to have reached both states of \"abiding\" and \"non-abiding nirvana\"."
        ],
        [
            "TEXT",
            "Kurt Cobain was the co-founder of the rock band Nirvana, along with Krist Novoselic and Aaron Burckhard."
        ],
        [
            "TABLE",
            "NIRVANAnet was founded in 1989."
        ]
    ]
}
\end{lstlisting}
\caption{Example of a refinement hallucination case (\system (GPT-3.5)). ``Dave Grohl'' is completely hallucinated.}
\label{lst:refinement_hallucination_1}
\end{figure*}

\begin{figure*}[ht]
\begin{lstlisting}[language=json]
{
    "idx": 249,
    "correct_hallucination": false,
    "question": "What is the voice type of the Bob Dylan?",
    "gold": "baritone",
    "answer_generated": "gravelly or nasal",
    "gold_sources": [
        "KB"
    ],
    "pred_sources": [],
    "evidences": [
        [
            "KB",
            "Wikidata says the answer to \"What is the voice type of the Bob Dylan?\" is: baritone."
        ],
        [
            "TEXT",
            "Bob Dylan's voice has been described as \"young and jeeringly cynical\" and \"broken\" as he aged."
        ],
        [
            "TEXT",
            "Bob Dylan's voice has received critical attention, with some describing it as \"a rusty voice\" and others comparing it to \"sand and glue\"."
        ]
    ]
}
\end{lstlisting}
\caption{Example of a refinement hallucination case}
\label{lst:refinement_hallucination_2}
\end{figure*}



\begin{figure*}[t]
\begin{lstlisting}[language=json]
messages = [
    {"role": "system", "content": "You are a helpful assistant."},
    {"role": "user", 
     "content": "You are a named entity recognition and entity disambiguation system. You are given a question and you need to list all entities in the question with a brief description for each entity. Each description should be max 10 words. Here are some examples:

     Question: what year lebron james came to the nba?
     Answer:
     1. LeBron James is American basketball player (born 1984)
     2. National Basketball Association is North American professional sports league 
    
     Question: what form of government was practiced in sparta?
     Answer:
     1. Sparta is city-state in ancient Greece
    
     Question: What is the genre of the tv series High Seas?
     Answer:
     1. High Seas is a Spanish television series

     Question: Which country did the TV series Coupling originate?
     Answer:
     1. Coupling is a British television series (2000-2004)

     Question: What year was M.O.V.E first formed?
     Answer:
     1. M.O.V.E is a Japanese musical group

     Question: What year was the inception of the soccer club Manchester United F.C.?
     Answer:
     1. Manchester United F.C. is association football club in Manchester, England

     Question: What is Russell Crowe's date of birth?
     Answer:
     1. Russell Crowe is New Zealand-born actor (born 1964)

     Question: what character did natalie portman play in star wars?
     Answer:
     1. natalie portman is Israeli-American actress and filmmaker
     2. star wars is epic space opera multimedia franchise created by George Lucas
     
     Question: what country is the grand bahama island in?
     Answer:
     1. Grand Bahama is island of the Bahamas
            
     Question: where are the nfl redskins from?
     Answer:
     1. Washington Commanders or Washington Redskins is American football team in the National Football League
            
     Question: what time zone am i in cleveland ohio?
     Answer:
     1. Cleveland is city in and county seat of Cuyahoga County, Ohio, United States
            
     Question: who is the prime minister of ethiopia?
     Answer:
     1. Ethiopia is country in the Horn of Africa},
     
    {"role": "user",
     "content": "List the entities and their descriptions for this question: 
     Question: {question} 
     Answer:"}
]
\end{lstlisting}
\caption{A shortened version of the prompt for GPT-3.5 to detect entity mentions and generate a description for each detected entity, as discussed in Section \ref{sec:knowledge_base}. The descriptions in the prompt are taken from the Wikidata description for detected entities. The actual prompt contains 13 more examples. The examples in the prompt are chosen to capture the diversity of domains and to instruct GPT-3.5 to detect more generic entities too.}
\label{tab:ned-prompt}
\end{figure*}

\begin{figure*}[ht]
\begin{lstlisting}[language=json]
{
    "question": "Who won the Oscars for the best actress in 1952?",
    "gold": "Vivien Leigh",
    "answer_generated": "Shirley Booth",
    "evidences": [
        [
            "KB",
            "Wikidata says the answer to \"Who won the Oscars for the best actress in 1952?\" is: ."
        ],
        [
            "TEXT",
            "Shirley Booth won the Academy Award for Best Actress in 1952."
        ]
    ]
}
\end{lstlisting}
\caption{A failure case after platinum evaluation. In this case, the intent of the question is ambiguous. The gold answer is Vivien Leigh, who won the 24th Academy Awards (held on March 20, 1952, honoring the films of 1951), and \system predicts Shirley Booth, who won The 25th Academy Awards (held on March 19, 1953, honoring the films of 1952).}
\label{lst:platinum_wrong_1}
\end{figure*}

\begin{figure*}[ht]
\begin{lstlisting}[language=json]
{
    "question": "Has Ericson Core started Career as music video director?",
    "gold": "Yes",
    "answer_generated": "No",
    "evidences": [
        [
            "KB",
            "Wikidata says the answer to \"Has Ericson Core started Career as music video director?\" is: no."
        ],
        [
            "TEXT",
            "Ericson Core started his career as a music video director."
        ]
    ]
}
\end{lstlisting}
\caption{A failure case after platinum evaluation. In this case, the Wikidata is giving an incorrect answer, due to semantic parsing errors.}
\label{lst:platinum_wrong_2}
\end{figure*}

\begin{figure*}[ht]
\begin{lstlisting}[language=json]
{
    "question": "Release year of the first Francisco de Robles book?",
    "gold": "1605",
    "answer_generated": "1585",
    "evidences": [
        [
            "KB",
            "Wikidata says the answer to \"Release year of the first Francisco de Robles book?\" is: ."
        ],
        [
            "TEXT",
            "Among the books published by Francisco de Robles, the first edition of \"Don Quixote\" was released in 1605."
        ],
        [
            "TABLE",
            "The first Francisco de Robles book, La Galatea, was released in the year 1585."
        ]
    ]
}
\end{lstlisting}
\caption{A failure case after platinum evaluation. The first book of Francisco de Robles is \textit{Don Quixote} released in 1605. The book \textit{La Galatea}, released in 1585, was published by Blas de Robles, father of Francisco de Robles. However, the entry of Blas de Robles, listed as the publisher in the Wikipedia infobox of \textit{La Galatea}, erroneously contains a hyperlink directing to the page of Francisco de Robles. This discrepancy led to a misinterpretation by \system, resulting in the incorrect identification of Francisco de Robles as the publisher of \textit{La Galatea}.}
\label{lst:platinum_wrong_3}
\end{figure*}

\begin{figure*}[ht]
\begin{lstlisting}[language=json]
{
    "question": "Is the player number of Bebe is 10?",
    "gold": "Yes",
    "answer_generated": "No",
    "evidences": [
        [
            "KB",
            "Wikidata says the answer to \"Is the player number of Bebe is 10?\" is: ."
        ],
        [
            "TABLE",
            "The player number of Bebe is 22."
        ]
    ]
}
\end{lstlisting}
\caption{A failure case after platinum evaluation. The player name, Bebé, was not correctly grounded in Wikidata, which resulted in an empty response. Table retriever retrieved information about a basketball player named Bebo instead of the football player Bebé.}
\label{lst:platinum_wrong_4}
\end{figure*}

\begin{figure*}[ht]
\begin{lstlisting}[language=json]
{
    "question": "Which island is home to Alyssa Cole's primary residence?",
    "gold": "Martinique",
    "answer_generated": "Information not available",
    "evidences": [
        [
            "KB",
            "Wikidata says the answer to \"Which island is home to Alyssa Cole's primary residence?\" is: ."
        ]
    ]
}
\end{lstlisting}
\caption{A failure case after platinum evaluation due to no retrieved evidence.}
\label{lst:platinum_wrong_5}
\end{figure*}

\begin{figure*}[ht]
\begin{lstlisting}[language=json]
{
    "question": "What is the original title of the novel The Alchemist?",
    "gold": "O Alquimista",
    "answer_generated": "\"O Alquimista\"",
}
\end{lstlisting}
\caption{Example where EM cannot handle correctly (format).}
\label{lst:em_cannot_handle_1}
\end{figure*}

\begin{figure*}[ht]
\begin{lstlisting}[language=json]
{
    "question": "Nirvana was founded by who?",
    "gold": "Kurt Cobain",
    "answer_generated": "Kurt Cobain and Krist Novoselic",
}
\end{lstlisting}
\caption{Example where EM cannot handle correctly (superset).}
\label{lst:em_cannot_handle_2}
\end{figure*}

\begin{figure*}[ht]
\begin{lstlisting}[language=json]
{
    "question": "What was Elton John's debut album?",
    "gold": "Goodbye Yellow Brick Road",
    "answer_generated": "Empty Sky",
}
\end{lstlisting}
\caption{Example where EM cannot handle correctly (gold answer wrong). ``Empty Sky'' is the correct answer here.}
\label{lst:em_cannot_handle_3}
\end{figure*}

\begin{figure*}[ht]
\begin{lstlisting}[language=json]
{
    "question": "What is the main cast name in the tv series Tribes of Europa?",
    "gold": "Emilio Sakraya",
    "answer_generated": "Henriette Confurius, Emilio Sakraya, and David Ali Rashed",
}
\end{lstlisting}
\caption{Example where EM cannot handle correctly (multiple correct answers).}
\label{lst:em_cannot_handle_4}
\end{figure*}

\begin{figure*}[ht]
\begin{lstlisting}[language=json]
{
    "question": "Who was the music of the movie \"The Social Network\"?",
    "gold": "Trent Reznor Atticus Ross",
    "answer_generated": "Trent Reznor and Atticus Ross",
}
\end{lstlisting}
\caption{Example where EM cannot handle correctly (paraphrasing).}
\label{lst:em_cannot_handle_5}
\end{figure*}

\end{document}